\documentclass[runningheads]{llncs}
\usepackage{makeidx}
\usepackage{graphicx}
\usepackage{cvpr-abbriv}
\usepackage{amsmath,amssymb} 
\usepackage{color}
\usepackage[ruled,vlined,linesnumbered,procnumbered,longend]{algorithm2e}
\usepackage{mathtools}

\DeclareMathOperator{\sign}{sign}
\usepackage{booktabs}
\usepackage{multirow}
\usepackage{array}

\usepackage[draft=false, pagebackref=false,breaklinks=true,colorlinks,bookmarks=false,pdftex]{hyperref}
\hypersetup{hidelinks,colorlinks=false,linkbordercolor={1 1 1}}
\usepackage[capitalise,nameinlink]{cleveref}
\let\parSym\S
\crefname{section}{\parSym}{\parSym\parSym}
\Crefname{section}{\parSym}{\parSym\parSym}

\graphicspath{{fig/}}

\def\Real{\mathbb{R}}
\def\Integer{\mathbb{Z}}

\def\O{\mathcal{O}}
\def\E{\mathcal{E}}
\def\S{\mathcal{S}}

\def\W{\mathcal{W}}
\makeatletter
\let\vaccent\v

\def\u{\textsc{u}}
\def\v{\textsc{v}}
\makeatother

\begin{document} \pagestyle{headings} \mainmatter

 \def\GCPR17SubNumber{32}

 \title{Scalable Full Flow with Learned Binary Descriptors}

 \titlerunning{Scalable Full Flow with Learned Binary Descriptors}
\authorrunning{G. Munda, A. Shekhovtsov, P. Kn\"obelreiter and T. Pock}
\author{Gottfried Munda$^1$ \quad Alexander Shekhovtsov$^2$ \quad Patrick Kn\"obelreiter$^1$ \\ Thomas Pock$^{1,3}$}
\institute{$^1$Institute of Computer Graphics and Vision, Graz University of Technology, Austria\\
  $^2$Czech Technical University in Prague\\
$^3$Center for Vision, Automation and Control, Austrian Institute of Technology }

\maketitle

\begin{abstract} 
We propose a method for large displacement optical flow in which local matching costs are learned by
a convolutional neural network (CNN) and a smoothness prior is imposed by a conditional random field (CRF). 
We tackle the computation- and memory-intensive operations on the 4D cost volume by a {\em
  min-projection} which reduces memory complexity from quadratic to linear and {\em binary descriptors} for efficient matching. 
This enables evaluation of the cost on the fly and allows to perform learning and CRF inference on
high resolution images without ever storing the 4D cost volume.
To address the problem of learning binary descriptors we propose a new hybrid learning scheme. 
In contrast to current state of the art approaches for learning binary CNNs we can compute the exact
non-zero gradient within our model.
We compare several methods for training binary descriptors and show results on public available benchmarks.
\end{abstract}

\section{Introduction}\label{sec:introduction}
Optical flow can be seen as an instance of the dense image matching problem, where the goal is to
find for each pixel its corresponding match in the other image. One fundamental question in the
dense matching problem is how to choose good \emph{descriptors} or \emph{features}. 
Data mining with convolutional neural networks (CNNs) has recently shown excellent results for
learning task-specific image
features, outperforming previous methods based on hand-crafted descriptors. 
One of the major difficulties in learning features for optical flow is the high
dimensionality of the cost function: Whereas in stereo, the full cost function can be represented as
a 3D volume, the matching cost in optical flow is a 4D volume. Especially
at high image resolutions, operations on the flow matching cost are expensive both in terms of memory requirements and
computation time. 

Our method avoids explicit storage of the full cost volume, both in the learning phase and during inference. 
This is achieved by a {\em splitting} (or {\em min-projection}) of the 4D cost into two
quasi-independent 3D volumes, corresponding to the $u$ and $v$ component of the flow.  
We then formulate CNN learning and CRF inference in this reduced setting. 
This achieves a space complexity linear in the size of the search range, similar to recent stereo
methods, which is a significant reduction compared to the quadratic complexity of the full 4D cost function.

Nevertheless, we still have to compute all entries of the 4D cost. This computational bottleneck can be optimized by
using binary descriptors, which give a theoretical speed-up factor of $32$. In practice, even larger
speed-up factors are attained, since binary descriptors need less memory bandwidth and also yield a better cache efficiency.
Consequently, we aim to incorporate a binarization step into the learning. 
We propose a novel hybrid learning scheme, where we circumvent the problem of hard nonlinearities having zero gradient. 
We show that our hybrid learning performs almost as well as a network without hard nonlinearities,
and much better than the previous state of the art in learning binary CNNs.

\section{Related Work}\label{sec:related-work}

In the past hand-crafted descriptors like SIFT, NCC, FAST
etc. have been used extensively with very good results, but recently CNN-based approaches 
\cite{Zbontar2015a,Luo2016} marked a paradigm shift in the field of image matching. To date all top performing methods
in the major stereo benchmarks rely heavily on features learned by CNNs. For optical flow, many
recent works still use engineered features \cite{Chen_2016_CVPR,Bailer2015}, presumably due to the difficulties the high
dimensional optical flow cost function poses for learning.
Only very recently we see a shift towards CNNs for learning descriptors
\cite{Gadot2016,Guney2016ACCV,Xu2017DCFlow}. Our work is most related to \cite{Xu2017DCFlow}, who 
construct the full 4D cost volume and run an adapted version of SGM on it. They perform learning and cost
volume optimization on $\tfrac{1}{3}$ of the original resolution and compress the cost function in
order to cope with the high memory consumption. Our method is memory-efficient thanks to the
dimensionality-reduction by the min-projection, and we outperform the reported runtime of
\cite{Xu2017DCFlow} by a factor of $10$.

Full flow with CRF~\cite{Chen_2016_CVPR} is a related inference method using
TRW-S~\cite{Kolmogorov-06-convergent-pami} with efficient distance transform
~\cite{FelsenszwalbHuttenlocherEfficientBeliefPropagation}. Its iterations have quadratic time 
and space complexity. 
In practice, this takes~20GB\footnote{Estimated for the cost volume size
  $341{\times}145{\times}160{\times}160$ based on numbers in~\cite{Chen_2016_CVPR} corresponding to 
  $\frac{1}{3}$ resolution of Sintel images.} of memory, and 10-30 sec. per iteration with a
parallel CPU implementation. We use the decomposed model~\cite{Shekhovtsov-Kovtun-Hlavac-08-CVIU}
with a better memory complexity and a faster parallel inference scheme based
on~\cite{Shekhovtsov2016}. 

Hand-crafted Binary Descriptors like Census have been shown to work well in a number of
applications, including image matching for stereo and flow	
~\cite{ranftl2014non,Ranftl2012,Trzcinski_2013_CVPR,Calonder_2010}. However, direct learning of binary descriptors is a
difficult task, since the hard thresholding function, $\sign(x)$, has gradient zero almost
everywhere. In the context of
Binary CNNs there are several approaches to train networks with binary
activations~\cite{Bengio2013} and even binary weights~\cite{Courbariaux2016,Rastegari2016a}. This is
known to give a considerable compression and speed-up at the price of a tolerable loss of
accuracy. To circumvent the problem of $\sign(x)$ having zero gradient a.e., surrogate gradients are
used. 
The simplest method, called {\em straight-through
  estimator}~\cite{Bengio2013} is to assume the derivative of $\sign(x)$ is 1, \ie, simply omit the
$\sign$ function in the gradient computation. This approach can be considered as the state of the
art, as it gives best results in~\cite{Bengio2013,Courbariaux2016,Rastegari2016a}. We show that in
the context of learning binary descriptors for the purpose of matching,
alternative strategies are possible which give better results.

\section{Method}\label{sec:model}
We define two models for optical flow: a local model, known as Winner-Takes-All (WTA) and a joint model, which uses CRF
inference. Both models use CNN descriptors, learned in~\cref{sec:learned-dataterm}. The joint model
has only few extra parameters that are fit 
separately and the inference is solved with a parallel method, see \cref{sec:crf}. For CNN learning,
we optimize the performance of the local model. While learning by optimizing the performance of the
joint model is possible \cite{KnobelreiterRSP16}, the resulting procedures are significantly more
difficult. 
\par
We assume color images $I^1,I^2\ :\ \Omega \to \mathbb{R}^3$, where $\Omega  = \{1,\dots
H\}\times\{1,\dots W\}$ is a set of pixels. Let $\mathcal{W}=\mathcal{S}\times\mathcal{S}$ be a
window of discrete 2D displacements, with $\mathcal{S}=\{-D/2, -D/2+1,\ldots,D/2-1 \}$ given by the
search window size $D$, an even number.
The flow $x\ :\ \Omega \to \mathcal{W}$ associates a displacement to each pixel $i\in\Omega$ so that
the displaced position of $i$ is given by $i+x_i \in \Integer^2$.  For convenience, we denote by
$x=(u,v)$, where $u$ and $v$ are mappings $\Omega \to \mathcal{S}$, the components of the flow in
horizontal and vertical directions, respectively.
The per-pixel {\em descriptors} $\phi(I;\theta)\ :\
\Omega\to\mathbb{R}^m$ are computed by a CNN with parameters $\theta$.
Let $\phi^1,\phi^2$ be descriptors of images $I^1$, $I^2$, respectively. The \emph{local matching
  cost} for a pixel $i\in\Omega$ and displacement $x_i\in\mathcal{W}$ is given by 
 \begin{equation}\label{local-cost}
   c_i(x_i) = \begin{cases}
     d(\phi^1_i, \phi^2_{i+x_i}) & {\rm if}\ i+x_i \in \Omega,\\
     c_{\rm outside} & {\rm otherwise},
   \end{cases}
 \end{equation}
where $d\ :\ \mathbb{R}^m\times\mathbb{R}^m\to\mathbb{R}$ is a distance function in
$\mathbb{R}^m$. ``Distance'' is used in a loose sense here, we will consider the negative\footnote{since we
  want to pose matching as a minimization problem} scalar product
$d(\phi^1,\phi^2)=-\langle\phi^1,\phi^2 \rangle$. 
We call
\begin{equation}\label{local-problem}
	\hat x_i \in \arg\min_{x_i \in \W} c_i(x_i)
\end{equation}
the \emph{local optical flow model}, which finds independently for each pixel $i$ a displacement
$x_i$ that optimizes the local matching cost. 
The \emph{joint optical flow model} finds the full flow field $x$ optimizing the coupled CRF energy cost:
\begin{equation}\label{joint-problem}
  \hat x \in \arg\min_{u,v \colon \Omega \to \S } \Big[ \sum_{i \in \Omega} c_i(u_i,v_i) +
  \sum_{i\sim j} w_{ij} (\rho (u_i-u_j) + \rho (v_i-v_j) ) \Big],
\end{equation}
where $i \sim j$ denotes a 4-connected pixel neighborhood, $w_{ij}$ are
contrast-sensitive weights, given by $w_{ij} =  \exp(- \frac{\alpha}{3} \sum_{c \in \{R,G,B\}}|I^1_{i,c} -I^1_{j,c}|)$ and $\rho
\colon \Real \to \Real$ is a robust penalty function 
shown in \cref{fig:CFR-flow}(a).
\par

\subsection{Learning Descriptors}
A common difficulty of models~\eqref{local-problem} and~\eqref{joint-problem} is that they need to
process the 4D cost~\eqref{local-cost}, which involves computing distances in $\Real^m$ per
entry. Storing such cost volume takes $\O(|\Omega|D^2)$ space and evaluating it $\O(|\Omega|D^2 m)$
time. We can reduce space complexity to $\mathcal{O}(|\Omega|D)$ by avoiding explicit storage of the 4D cost function.
This facilitates memory-efficient end-to-end training on high resolution images, without a patch
sampling step \cite{Xu2017DCFlow,Luo2016}.
Towards this end we write the local optical flow model \eqref{local-problem} in the following way
\begin{subequations}\label{hat-u-v}
\begin{align}
\label{hat-u}
\hat u_i &\in \arg \min_{u_i} c^\u_i(u_i),\ \ \mbox{where} \ \ \
c^\u_i(u_i)  = \min_{v_i} c_i(u_i, v_i);\\
\label{hat-v}
\hat v_i &\in \arg \min_{v_i} c^\v_i(v_i),\ \ \mbox{where} \ \ \
c^\v_i(v_i)  = \min_{u_i} c_i(u_i, v_i).
\end{align}
\end{subequations}
The inner step in~\eqref{hat-u} and~\eqref{hat-v}, called {\em min-projection}, minimizes out one
component of the flow vector. This can be interpreted as a decoupling of the full 4D flow problem
into two simpler quasi-independent 3D problems on the reduced cost volumes $c^\u, c^\v$. 
Assuming the minimizer of~\eqref{local-problem} is unique, \eqref{hat-u} and \eqref{hat-v}
find the same solution as the original problem \eqref{local-problem}. Using this representation, CNN
learning can be implemented within existing frameworks. We point out that this
approach has the same space complexity as recent methods for learning stereo matching, since we only
need to store the 3D cost volumes $c^\u$ and $c^\v$. 
 As an illustrative example consider an image with size $1024 \times 436$ and a search range of
 256. In this setting the full 4D cost function takes roughly $108$\,GB whereas our splitting
 consumes only $0.8$\,GB.

\subsubsection{Network}\label{sec:learned-dataterm}
\begin{figure}[t]
  \centering
  \def\svgwidth{0.99\columnwidth}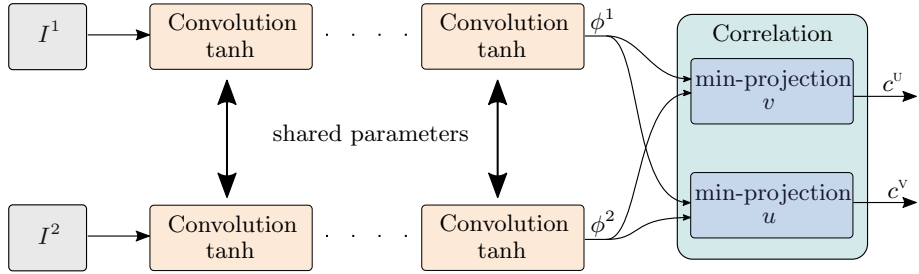
  \caption{Network architecture: A number of convolutional layers with
    shared parameters computes feature vectors $\phi^1,\phi^2$ for every pixel. These feature
    vectors are cross-fed into a correlation layer, that computes local matching costs in $u$ and $v$
    direction by minimizing out the other direction. The result are two quasi-independent
    cost volumes for the $u$ and $v$ component of the flow.}
  \label{fig:network}
\end{figure}

\cref{fig:network} shows the network diagram of the local flow model ~\cref{local-problem}.
The structure is similar to the recent methods proposed for learning stereo matching
\cite{Luo2016,Zbontar2015a,Chen2015,KnobelreiterRSP16}. It is a siamese network consisting of two
convolutional branches with shared parameters, followed by a correlation layer.
The filter size of the convolutions is $3\times 3$ for the first layer and $2\times 2$ for all other
layers. The $\tanh$ nonlinearity keeps feature values in a defined range, which works well with the
scalar product as distance function. We do not use striding or pooling. The last convolutional layer
uses 64 filter channels, all other layers have 96 channels. This fixes the dimensionality of the
distance space to $m=64$.

\textbf{Loss}
Given the groundtruth flow field $(u^*,v^*)$, we pose the learning objective as follows: we define a
probabilistic softmax model of the local prediction $u_i$ (resp. $v_i$) as $p(u_i) \propto \exp(-c_i^\u(u_i))$,
then we consider a naive model $p(u,v) = \prod_{i}p(u_i)p(v_i)$ and apply the maximum likelihood
criterion. The negative log likelihood is given by
\begin{equation}
  \label{eq:cnn-surrogate-loss}
  L(u,v) = -\sum_{i\in\Omega} \big[ \log p(u_i^*) +\log p(v_i^*) \big].
\end{equation}
This is equivalent to cross-entropy loss with the target distribution concentrated at the single
point $(u_i^*,v_i^*)$ for each $i$. Variants of the cross-entropy loss, where the target
distribution is spread around the ground truth point $(u_i^*,v_i^*)$ are also used in the
literature~\cite{Luo2016} and can be easily incorporated.
\subsubsection{Learning Quantized Descriptors}
The computational bottleneck in scheme~\eqref{hat-u-v} is computing the min-projections, with time complexity $\O(|\Omega|D^2m)$. 
This operation arises during the learning as well as in the CRF inference step, where it corresponds
to the message exchange in the dual decomposition. It is therefore desirable 
to accelerate this step.
We achieve a significant speed-up by quantizing the descriptors and evaluating the Hamming distance
of binary descriptors. 

Let us  define the quantization: we call $\bar \phi = \sign(\phi)$ the {\em quantized
  descriptor field}. 
The distance between quantized descriptors is given by $d(\bar\phi^1,\bar\phi^2) = -\langle \bar\phi^1,
\bar\phi^2 \rangle = 2\mathcal{H}(\bar\phi^1,\bar\phi^2)-m$, equivalent to the Hamming distance $\mathcal{H}(\cdot,\cdot)$
up to a scaling and an offset.
 Let the quantized cost function be denoted
$\bar c_i(x_i)$, defined similar to~\eqref{local-cost}. We can then compute quantized min-projections $\bar c^\u$, $\bar c^\v$.
 
However, learning model~\eqref{local-problem} with quantized descriptors is difficult due to the
gradient of the sign function being zero almost everywhere.  
We introduce a new technique specific to the matching problem and compare it to the baseline
 method that uses the straight-through estimator of the gradient~\cite{Bengio2013}. 
Consider the following variants of the model~\eqref{hat-u} 
\begin{align}
\tag{FQ}\label{FQ}
&\hat u_i \in \arg \min_{u_i} c_i(u_i,\hat v_i(u_i)),\ \ \mbox{where} \ \ \
\hat v_i(u_i) \in \arg\min_{v_i} \bar c_i(u_i, v_i);\\
\tag{QQ}\label{QQ}
&\hat u_i \in \arg \min_{u_i} \bar c_i(u_i,\hat v_i(u_i)),\ \ \mbox{where} \ \ \
\hat v_i(u_i) \in \arg\min_{v_i} \bar c_i(u_i, v_i).
\end{align}
The respective variants of~\eqref{hat-v} are symmetric.
The second letter in the naming scheme indicates whether the inner problem, \ie, the min-projection
step, is performed on (Q)uantized or (F)ull cost, whereas the first letter refers to the outer
problem on the smaller 3D cost volume. The initial model~\eqref{hat-u} is thus also denoted as FF model. While
models FF and QQ correspond, up to non-uniqueness of solutions, to the joint minimum in $(u_i,v_i)$
of the cost $c$ and $\bar c$ respectively, the model FQ is a mixed one. This hybrid model is
interesting because minimization in $v_i$ can be computed efficiently on the binarized cost with
Hamming distance, and the minimization in $u_i$ has a non-zero gradient in $c^{\u}$.
We thus consider the model FQ as an efficient variant of the local optical flow model~\eqref{local-problem}. 
In addition, it is a good learning proxy for the model QQ: 
Let $\hat u_i = \arg\min_{u_i} c_i(u_i,\hat v_i(u_i))$ be a minimizer of the outer
problem FQ. Then the derivative of FQ is defined by the indicator of the pair $(\hat u_i, 
\hat v_i(u_i))$.
This is the same as the derivative of FF, except that $\hat v_i(u_i)$ is computed differently.
Learning the model QQ involves a hard quantization step, and we apply the straight-through estimator
to compute a gradient. Note that the exact gradient for the model FQ can be computed at
approximately the same reduced computational cost as the straight-through gradient in the model QQ.
\par

\subsection{CRF}\label{sec:crf}
\begin{figure}
\centering
\begin{tabular}{ccc}
\begin{tabular}{c}\includegraphics[width=0.15\linewidth]{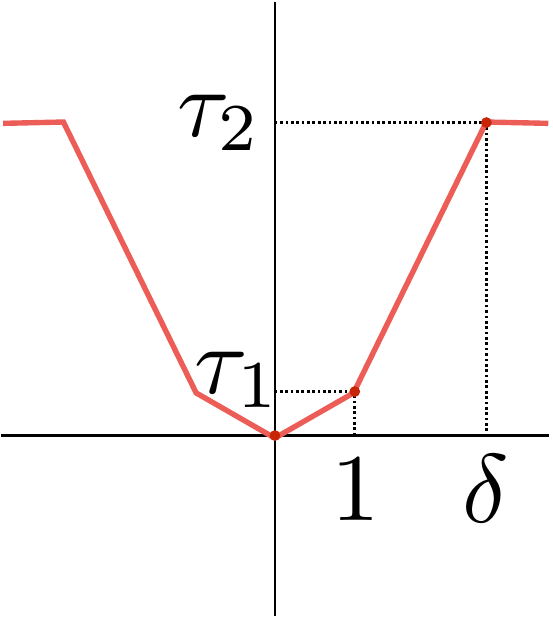}\end{tabular}&\ \
\begin{tabular}{c}\includegraphics[width=0.5\linewidth]{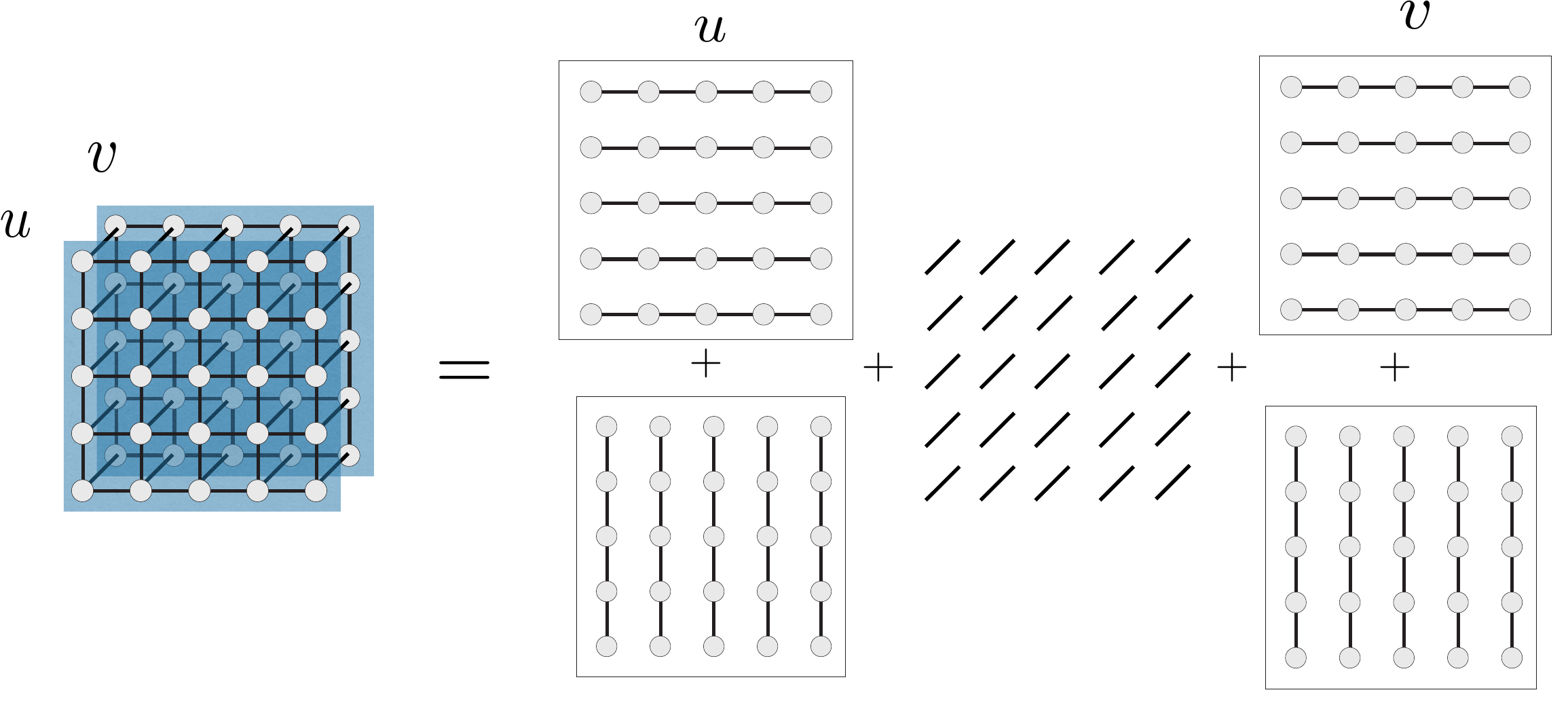}\end{tabular}&\ \
\begin{tabular}{c}\includegraphics[height=0.15\linewidth]{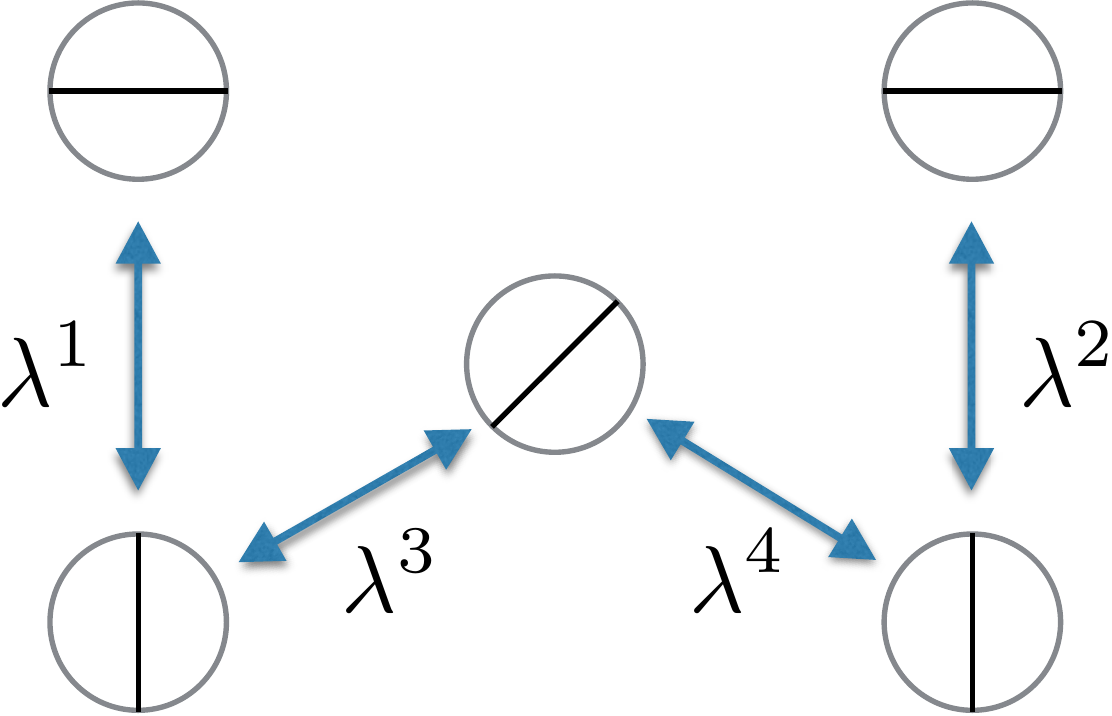}\end{tabular}\\
(a) & (b) & (c)
\end{tabular}
\caption{Building blocks of the CRF. (a) Robust pairwise function $\rho$. (b) Decomposition of the
  pairwise CRF into 5 subproblems. (c) Lagrange multipliers in the dual corresponding to equality
  constraints between the subproblems. They act as offsets of unary costs between subproblems,
  increasing on one side of the arrow and decreasing on the other. }\label{fig:CFR-flow} 
\end{figure}

The baseline model, which we call {\em product} model, has $|\Omega|$ variables $x_i$ with the state
space $S\times S$. It has been observed in~\cite{FelsenszwalbHuttenlocherEfficientBeliefPropagation}
that max-product message passing in the CRF~\eqref{joint-problem} can be computed in time
$\mathcal{O}(D^2)$ per variable for separable interactions using a fast distance transform. However,
storing the messages for a 4-connected graph requires $\mathcal{O}(|\Omega|D^2)$ memory. Although
such an approach was shown feasible even for large displacement optical flow~\cite{Chen_2016_CVPR},
we argue that a more compact {\em decomposed} model~\cite{Shekhovtsov-Kovtun-Hlavac-08-CVIU} gives
comparable results and is much faster in practice.
The decomposed model is constructed by observing that the regularization in~\eqref{joint-problem} is
separable over $u$ and $v$. Then the energy~\eqref{joint-problem} can be represented as a CRF with
$2|\Omega|$ variables $u_i,v_i$ with the following pairwise terms: The in-plane term
$w_{ij}\rho(u_i-u_j)$ and the cross-plane term $c(u_i,v_i)$, forming the graph shown
in~\cref{fig:CFR-flow}(b). In this formulation there are no unary terms, since costs $c_i$ are
interpreted as pairwise terms.
The resulting linear programming (LP) dual is more economical, because it has only
$\mathcal{O}(|\Omega| D)$ variables. The message passing for edges inside planes and across planes
has complexity $\mathcal{O}(|\Omega| D)$ and $O(|\Omega| D^2)$, respectively.

We apply the parallel inference method~\cite{Shekhovtsov2016} to the dual of the decomposed
model~\cite{Shekhovtsov-Kovtun-Hlavac-08-CVIU} (see~\cref{fig:CFR-flow}(b)). 
Although different dual decompositions reach different objective values in a fixed number of iterations, 
it is known that all decompositions with trees covering the graph are equivalent in the optimal
value~\cite{Wainwright-MAP}. The decomposition in~\cref{fig:CFR-flow}(b) is into horizontal and
vertical chains in each of the $u$- and $v$- planes plus a subproblem containing all cross-layer
edges. We introduce Lagrange multipliers $\lambda=(\lambda^k \in\Real^{\Omega \times \S}\mid
k=1,2,3,4)$ enforcing equality constraints between the subproblems as shown
in~\cref{fig:CFR-flow}(c). The Lagrange multipliers $\lambda^k$ are identified with modular
functions $\lambda^k \colon \S^\Omega \to \Real \colon  u \mapsto \sum_{i}\lambda^k_i(u_i)$. 
Let us also introduce shorthands for the sum of pairwise terms over horizontal chains $f^{\rm
  h}\colon \S^\Omega \to \Real\colon u \mapsto \sum_{ij\in\E^{\rm h}}w_{ij}\rho(u_i-u_j)$, and a
symmetric definition $f^{\rm v}$ for the sum over the vertical chains.
The lower bound $\Psi(\lambda)$ corresponding to the decomposition in~\cref{fig:CFR-flow}(c) is given by:
\begin{subequations}
\begin{align}\label{LB-total}
\Psi(\lambda) = \ &\Psi^1(\lambda) + \Psi^2(\lambda) + \Psi^3(\lambda), \mbox{\ where} \\
\label{LB-plane-u}
& \Psi^1(\lambda) = \min\limits_{u}\big[ (\lambda^{1}+\lambda^{3})(u) + f^{\rm h}(u) \big] + 
\min\limits_{u}\big[ -\lambda^{1}(u) + f^{\rm v}(u) \big];\\
\label{LB-plane-v}
& \Psi^2(\lambda) = \min\limits_{v}\big[ (\lambda^{2}+\lambda^{4})(v) + f^{\rm h}(v) \big] + 
\min\limits_{v}\big[ -\lambda^{2}(v) + f^{\rm v}(v) \big];\\
\label{LB-uv}
& \Psi^3(\lambda) = \sum_{i}\min_{u_i, v_i} \big[ c_i(u_i,v_i) -\lambda_i^{3}(u_i)-\lambda_i^{4}(v_i) \big].
\end{align}
\end{subequations}
Our Lagrangian dual to~\eqref{joint-problem} is to maximize $\Psi(\lambda)$ in $\lambda$, which
enforces consistency between minimizers of the subproblems. The general theory~\cite{Wainwright-MAP}
applies, in particular, when the minimizers of all subproblems are consistent they form a global
minimizer.
In~\eqref{LB-plane-u}, there is a sum of horizontal and vertical chain subproblems in the
$u$-plane. When $\lambda^3$ is fixed, $\Psi^1(\lambda)$ is the lower bound corresponding to the
relaxation of the energy in $u$ with the unary terms given by $\lambda^3$. It can be interpreted as
a stereo-like problem with 1D labels $u$. Similarly, $\Psi^2(\lambda)$ is a lower bound for the
$v$-plane with unary terms $\lambda^4$. Subproblem $\Psi^3(\lambda)$ is simple, it contains both
variables $u,v$ but the minimization decouples over individual pairs $(u_i,v_i)$. It connects the
two stereo-like problems through the 4D cost volume $c$.

Updating messages inside planes can be done at a different rate than across
planes. The optimal rate for fast convergence depends on the time complexity of the message
updates. \cite{Shekhovtsov-Kovtun-Hlavac-08-CVIU} reported an optimal rate of updating in-plane
messages 5 times as often using the TRW-S solver~\cite{Kolmogorov-06-convergent-pami}. The
decomposition~\eqref{LB-total} facilitates this kind of strategy and allows to use the
implementation~\cite{Shekhovtsov2016} designed for stereo-like problems.  
We therefore use the dual
solver~\cite{Shekhovtsov2016}, denoted Dual Minorize-Maximize (DMM) to perform in-plane
updates. When applied to the problem of maximizing $\Psi^1(\lambda)$ in $\lambda^1$, it has the
following properties: a) the bound $\Psi^1(\lambda)$ does not decrease and b) it computes a modular
minorant $s$ such that $s(u) \leq \lambda^{3}(u) + f^{\rm h}(u) + f^{\rm v}(u)$ for all $u$ and
$\Psi^1(\lambda) = \sum_{i} \min_{u_i} s_i(u_i)$. The modular minorant $s$ is an excess of costs,
called {\em slacks}, which can be subtracted from $\lambda^3$ while keeping $\Psi^1(\lambda)$
non-negative. 
The associated update of the $u$-plane can be denoted as
\begin{subequations}
\begin{eqnarray}\label{dmm-update-u}
(\lambda^1, s) && := {\rm DMM}(\lambda^1,\lambda^3,f^{\rm h},f^{\rm v}),\\
\lambda^3 && := \lambda^3 - s.
\end{eqnarray}
\end{subequations}
The slack $s$ is then passed to the $v$ plane by the following updates, \ie, message passing $u \rightarrow v$:
\begin{equation}\label{u-v-msg}
\lambda_i^{4}(v_i) := \lambda_i^{4}(v_i) + \min_{u_i} \big[ c_i(u_i,v_i) -\lambda_i^{3}(u_i) \big].
\end{equation}
The minimization~\eqref{u-v-msg} 
has time complexity $\mathcal{O}(|\Omega|D^2)$, assuming the 4D costs $c_i$ are available in memory.
As discussed above, we can compute the costs $c_i$ efficiently on the fly 
and avoid $\mathcal{O}(|\Omega|D^2)$ storage.
The update $v\rightarrow u$ is symmetric to~\eqref{dmm-update-u}: 
\begin{equation}\label{v-u-msg}
\lambda_i^{3}(u_i) := \lambda_i^{3}(u_i) + \min_{v_i} \big[ c_i(u_i,v_i) -\lambda_i^{4}(v_i) \big].
\end{equation}
\par
The complete method is summarized in~\cref{A:coupled}. 
It starts from collecting the slacks in the $u$-plane. When initialized with $\lambda=0$, the
update~\eqref{v-u-msg} simplifies to $\lambda^3_i(u_i) = \min_{v_u} c_i(u_i,v_i)$, \ie, it is
exactly matching to the min-projection $c^\u$~\eqref{hat-u-v}. The problem solved with DMM
in~\cref{step-dmm-u} in the first iteration is a stereo-like problem with cost $c^\u$. The dual
solution redistributes the costs and determines which values of $u$ are worse than others, and
expresses this cost offset in $\lambda^3$ as specified in~\eqref{dmm-update-u}. The optimization of
the $v$-plane then continues with some information of good solutions for $u$ propagated via the cost
offsets using~\eqref{u-v-msg}. 

\begin{algorithm}[tb]
\KwIn{Cost volume $c$\;}
\KwOut{Dual point $\lambda$ optimizing $\Psi(\lambda)$\;}
Initialize $\lambda :=0$\;
\For{$t=1,\dots ,\mbox{\texttt{\upshape it\_outer}}$}{
Perform the following updates:\\
\makebox[4em][r]{$v \rightarrow u$}: pass slacks to $u$-plane by~\eqref{v-u-msg}, changes $\lambda^3$\;
\makebox[4em][r]{$u$-plane}: DMM with \texttt{it\_inner} iterations for $u$-plane~\eqref{dmm-update-u}, changes $\lambda^1$, $\lambda^3$ \label{step-dmm-u}\;
\makebox[4em][r]{$u \rightarrow v$}: pass slacks to $v$-plane by~\eqref{u-v-msg}, changes $\lambda^4$\;
\makebox[4em][r]{$v$-plane}: DMM with \texttt{it\_inner} iterations for $v$-plane, changes $\lambda^2$, $\lambda^4$\;
}
\caption{Flow CRF Optimization\label{A:coupled}}
\end{algorithm}

\section{Evaluation}
\label{sec:evaluation}
We compare different variants of our own model on the Sintel optical flow dataset~\cite{Butler2012}.
In total the benchmark consists of 1064 training images and 564 test images. For CNN learning we use
a subset of $20\%$ of the training images, sampled evenly from all available scenes. For evaluation,
we use a subset of $40\%$ of the training images. \newline \newline
\textbf{Comparison of our models}
To investigate the performance of our model, we conduct the following experiments: 
First, we investigate the influence of the size of the CNN,  and second we
investigate the effect of quantizing the learned features. 
Additionally, we evaluate both the WTA solution \eqref{local-problem}, and the
CRF model \eqref{joint-problem}. To assess the effect of quantization, we evaluate the local flow
model a) as it was trained, and b) QQ, \ie, with quantized descriptors both in the min-projection
step as well as in the outer problem on $c^\u, c^\v$ respectively. 
In CRF inference the updates \eqref{u-v-msg} and \eqref{v-u-msg} amount to solving a min-projection step with additional cost
offsets. F and Q indicate how this min-projection step is computed.
CRF parameters are fixed at $\alpha=8.5,~\eqref{joint-problem},\ \tau_1=0.25,\ \tau_2=25$
(\cref{fig:CFR-flow}) for all experiments and we run 8 inner and 5 outer iterations.
\cref{tab:modelComparison} summarizes the comparison of different variants of our model. 
\begin{table}[t]
  \caption{Comparison of our models on a representative validation set at scale $\tfrac{1}{2}$. We
    present the end-point-error (EPE) for non-occluded (noc) and all pixels on Sintel clean. }
\setlength{\tabcolsep}{5pt}
\centering
\begin{tabular}{cc||cc|cc}
\multicolumn{2}{c}{} & \multicolumn{2}{c}{\textbf{Local Flow Model (WTA)}} & \multicolumn{2}{c}{\textbf{CRF}}\\
\toprule
\textbf{Train} & \textbf{\#Layers} & \textbf{as trained} & \textbf{QQ} & \textbf{F} & \textbf{Q} \\
 & & noc (all) & noc (all) & noc (all) & noc (all)\\
\midrule
\multirow{3}{*}{FF} & 5 & 5.25 (10.38) & 10.45 (15.67) & 1.58 (4.48) & 1.64 (4.87)\\
& 7 &  4.72 (10.04) & 9.43 (14.93) & 1.53 (4.32) & 1.61 (4.70)\\
& 9 &  --\footnotemark[1] & --\footnotemark[1] & --\footnotemark[1] & --\footnotemark[1]\\
\midrule
\multirow{3}{*}{FQ} 
& 5 & 6.15 (11.36) & 11.43 (16.78) & --\footnotemark[2]  & 1.63 (4.62) \\
& 7 & 5.62 (10.98) & 10.15 (15.70) & --\footnotemark[2] & 1.65 (4.62) \\
& 9 & 5.62 (11.13) & 9.87 (15.52) & --\footnotemark[2] & 1.64 (4.69) \\
\midrule
  \multirow{3}{*}{QQ} 
  & 5 & same as QQ  & 9.63 (14.80) & --\footnotemark[2] &  1.72 (4.91)  \\
  & 7 & same as QQ & 9.75 (15.23) & --\footnotemark[2] & 1.66 (4.78)  \\
  & 9 & same as QQ & 9.72 (15.31) & --\footnotemark[2] & 1.72 (4.85)  \\
\bottomrule
\end{tabular}
\label{tab:modelComparison}
\end{table}
\footnotetext[1]{{\footnotesize Omitted due to very long training time.}}
\footnotetext[2]{{\footnotesize Not applicable.}}
We see that the WTA solution of model FQ performs
similarly to FF, while being much faster to train and evaluate. In particular, model FQ performs
better than QQ, which was trained with the straight through estimator of the gradient. If we
switch to QQ for evaluation, we see a drop in performance for models FF and FQ. This is to be
expected, because we now evaluate costs differently than during training.
Interestingly, our joint model yields similar performance regardless whether we use F or Q for
computing the costs. 

\subsubsection{Runtime}
\begin{figure}[ht]
\centering
\includegraphics[width=0.77\textwidth]{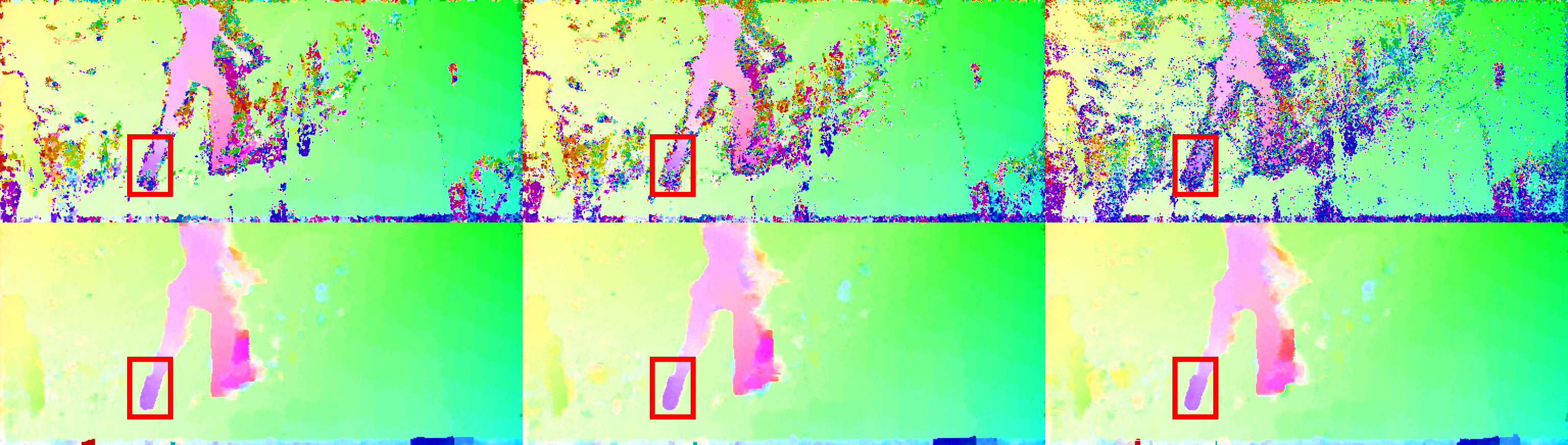}
\includegraphics[width=0.22\textwidth]{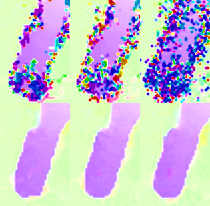}
\caption{Sample output of our method. Left figure, top row shows the WTA solution of a 7-layer network for {\em FF, FQ, QQ} training. The bottom row shows results of the same network with CRF inference. The right part shows the highlighted region enlarged.}
\label{fig:results}
\end{figure}

The main reason for quantizing the descriptors is speed. 
In CRF inference, we need to compute the min-projection on the 4D cost function twice per outer
iteration, see Alg. \ref{A:coupled}. 
We show an exact breakdown of the timings for $D=128$ on full resolution images in
\cref{tab:timings}, computed on a Intel i7 6700K and a Nvidia Titan X.
\begin{table}[t]
\smaller
\begin{minipage}{0.66\textwidth}
\centering
\caption{Timings of the building blocks (seconds).}
\begin{tabular}{cccc}
\toprule
{\textbf{Method}} & {\textbf{Feature Extraction}}\ \ \ & {\textbf{WTA}}\ \ \ & {\textbf{Full Model}} \\
\midrule
  FF & 0.04 -- 0.08 & 4.25 & 24.8\\ 
  FQ & 0.04 -- 0.08 & 1.82 & -\\ 
  QQ & 0.04 -- 0.08 & 0.07 & 3.2\\ 
\midrule
\cite{Xu2017DCFlow} ($\tfrac{1}{3}$ res.)      &  0.02 & 0.06  & 3.4 \\ 
QQ ($\tfrac{1}{3}$ res.)      & 0.004 -- 0.008 & 0.007 & 0.32\\
\bottomrule
\end{tabular}
\end{minipage}
\begin{minipage}{0.33\textwidth}
\caption{Comparison on the Sintel clean test set.}
\begin{tabular}{l|>{\centering\arraybackslash}p{1.1cm} >{\centering\arraybackslash}p{1.1cm} >{\centering\arraybackslash}p{1.1cm}|>{\centering\arraybackslash}p{1.1cm} >{\centering\arraybackslash}p{1.1cm} >{\centering\arraybackslash}p{1.1cm}|ccc}
\toprule
\textbf{Method} & \textbf{noc} & \textbf{all} \\
\midrule
EpicFlow~\cite{Revaud2015} & 1.360 & 4.115 \\
FullFlow~\cite{Chen_2016_CVPR} & 1.296 & 3.601\\
FlowFields~\cite{Bailer2015} & 1.056 & 3.748 \\
DCFlow~\cite{Xu2017DCFlow} & 1.103 & 3.537 \\
Ours QQ & 2.470 & 8.972\\
\bottomrule
\end{tabular}
\label{tab:timings}
\end{minipage}
\end{table}
The column WTA refers to computing the solution of the local model on the cost volumes $c^\u,c^\v$, see
\cref{hat-u-v}. Full model is the CRF inference, see \cref{sec:crf}. We see that we can reach a
significant speed-up by using binary descriptors and Hamming distance for computing intensive
calculations. For comparison, we also report the runtime of \cite{Xu2017DCFlow}, who, at the time of
writing, report the fastest execution time on Sintel. We point out
that our CRF inference on full resolution images takes about the same time as their
method, which constructs and optimizes the cost function at $\tfrac{1}{3}$ resolution. \newline \newline
\textbf{Test performance}
We compare our method on the Sintel clean images. 
In contrast to the other methods we do not use a sophisticated post-processing pipeline, because the
main focus of this work is to show that learning and inference on high resolution images is
feasible.  Therefore we cannot compete with the highly tuned methods. 
\cref{fig:results} shows that we are able to recover fine details, but since we do not employ a
forward-backward check and local planar inpainting we make large errors in occluded regions. 

\section{Conclusion}
We showed that both learning and CRF inference of the optical flow cost function on high resolution
images is tractable. We circumvent the excessive memory requirements of the full 4D cost volume by
a min-projection. This reduces the space complexity from quadratic to linear in the search range. To
efficiently compute the cost function, we learn binary descriptors with a new hybrid learning
scheme, that outperforms the previous state-of-the-art straight-through estimator of the gradient.
{
  \footnotesize
  \paragraph{Acknowledgements}
We acknowledge grant support from Toyota Motor Europe HS, the ERC starting grant ”HOMOVIS”
  No. 640156 and the research initiative Intelligent
Vision Austria with funding from the AIT and the Austrian Federal
Ministry of Science, Research and Economy HRSM programme
(BGBl. II Nr. 292/2012) 
}

\let\v\vaccent
\bibliographystyle{splncs03}
\bibliography{bib/all}

\end{document}